\documentclass{article}
\usepackage{amsmath,graphicx,amssymb}
\usepackage{xcolor}
\usepackage{graphics}
\usepackage{float}
\usepackage{multirow}
\usepackage{comment}
\usepackage{breqn}
\DeclareMathOperator*{\argmax}{argmax} 
\usepackage{enumitem}
\usepackage{adjustbox}
\usepackage{fancyhdr,lipsum}
\usepackage[preprint]{spconf}
\usepackage[top=30mm, bottom=20mm, left=18mm, right=18mm]{geometry}

\title{Object-centric and memory-guided normality reconstruction for video anomaly detection}

\name{\adjustbox{max width=\textwidth}{{Khalil Bergaoui$^{1 \ *}$, Yassine Naji$^{1,2}$ \sthanks{Equal contributions}, Aleksandr Setkov$^{1}$, Angélique Loesch$^{1}$, Michèle Gouiffès$^{2}$, Romaric Audigier$^{1}$}}}
\address{$^{1}$Université Paris-Saclay, CEA, List, 91120, Palaiseau, France \\ $^{2}$Université Paris-Saclay, CNRS, LISN, 91400, Orsay, France}

\begin{document}
%
\maketitle
\begin{abstract}

This paper addresses the anomaly detection problem for videosurveillance. Due to the inherent rarity and heterogeneity of abnormal events, this problem is tackled from a normality modeling perspective, where our model learns object-centric normal patterns without seeing anomalous samples during training. Our main contributions consist in coupling object-level action features with a cosine distance-based anomaly estimation function. We therefore extend previous methods by introducing explicit geometric constraints to the mainstream reconstruction-based strategy. Our framework leverages both appearance and motion information to learn object-level behavior and captures prototypical patterns within a memory module.  Experiments on several well-known datasets demonstrate the effectiveness of our method as it outperforms current state-of-the-art on most relevant spatio-temporal evaluation metrics.

\end{abstract}
\section{Introduction}
\label{sec:intro}

Video Anomaly Detection (VAD) is an open research problem which consists in detecting rare occurrences of abnormal events. This is a challenging problem due to two main reasons. 
Although anomalous events are generally defined as rare occurrences that deviate from normal patterns observed in familiar events \cite{ramachandra2020survey}, this definition does not differentiate anomalous events from rare normal ones.
Secondly, abnormal events are inherently more difficult to collect and to learn, due to their few occurrences and the multiplicity of their nature. For these reasons, the VAD problem is often viewed within the \emph{one-class}\textbf{} paradigm \cite{pang20review}.

In a pioneering work~\cite{liu2018ano_pred}, a model is trained to predict future ``normal'' frame, and 
anomalies are viewed as inaccurate predictions. 
The recent method~\cite{ssmt} combines
multiple proxy tasks (e.g. arrow of time prediction)  to characterize anomalous events. Other approaches 
quantify the deviation from learned normal patterns including distance-based \cite{dummyae, ionescu_narrowclusters,streetscene,siamese,any-shot} and reconstruction-based methods \cite{app-mo,memae,mnad,ssmt,baf}. 
Although 
they have been empirically shown to attain impressive performance levels on current standard benchmark datasets, the used strategy is not always in line with the nature of anomalous event detection. In fact, as pointed out in \cite{ssmt} , a car stopped in a pedestrian area should be labeled as an anomaly, yet the car is trivial to reconstruct (at a pixel-wise level) in a future frame, since it is still standing. 
This example shows that pixel-wise reconstruction error is suboptimal for anomaly detection. 
The recent work ~\cite{baf} addressed this challenge by learning object-level patterns of normal appearance and motion by training a discriminator network to classify (normal {\it vs.} abnormal samples) given pairs of reconstruction error maps. Despite being the current state of the art, this method includes out-of-domain observations during training, introducing a bias to the normality modeling. Instead, we propose to tackle these challenges by extending the mainstream reconstruction assumption on which most state-of-the-art methods \cite{liu2018ano_pred,memae,mnad,baf} are implicitly based: \textit{Given a normality model, normal observations are easier to reconstruct from a low-dimensional representation than abnormal observations}. We propose to add geometric constraints in the reconstruction space in order to further narrow down this assumption to be more in line with the anomaly detection task. Different from prior works, we combine a cosine distance-based anomaly estimation function with pretrained object-level features. Additionally, we propose to constrain our model to reconstruct independent motion and appearance features from a single embedding space. This way, our network has fewer degrees of freedom to perform the training task, 
which is in line with the aforementioned reconstruction assumption. Following \cite{dummyae,any-shot,cl_doshi20,ssmt,baf} 
we apply an object detector allowing to localize anomalies at the object level, which is semantically more relevant than at the pixel-level. Similarly to \cite{memae,mnad}, we incorporate a memory block in our framework in order to 
 model diverse normality patterns. In summary, our contributions are:  

\begin{figure*}
	\centering
	\includegraphics[width=0.9\textwidth]{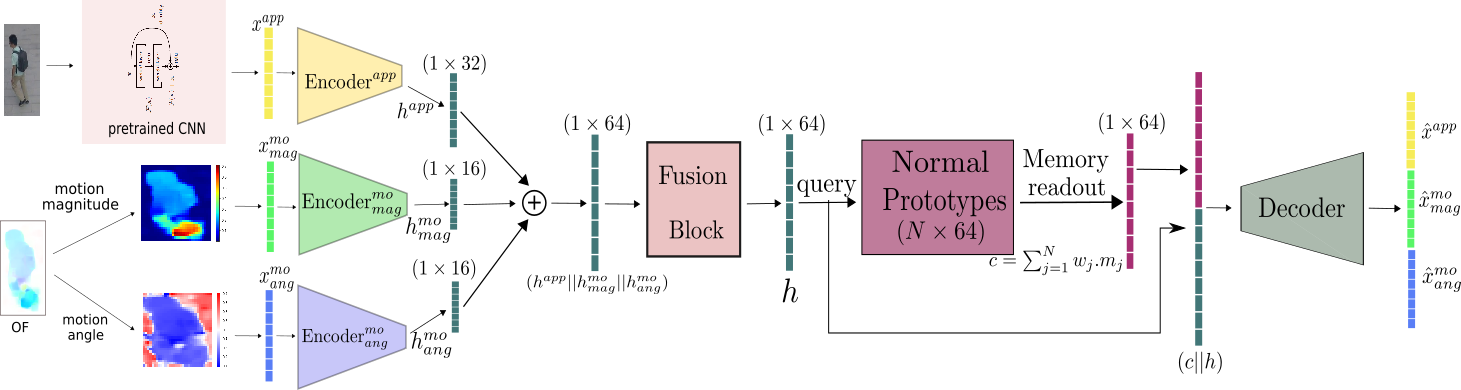}
	\caption{ Overview of OMAE. At the preprocessing step, object bounding boxes and Optical Flow (OF) are computed. Object appearance features, extracted using a pretrained CNN, motion magnitude and angle maps are fed to the corresponding autoencoders. The encoded representations are fused ($h$) and sent as a query to the memory module which fetches similar memory items. Their linear combination concatenated with $h$ is sent to the decoder to obtain object appearance and motion reconstructions. The anomaly score is defined based on the dissimilarity between the input features and their reconstructions as well the dissimilarity between the query $h$ and its neighbors.}
	\label{fig:omae_figure}
\end{figure*}

 $\bullet$ Imposing geometric constraints in the reconstruction space using cosine distance.
 
 $\bullet$ Introducing object-level action prototypical features.

  $\bullet$ State-of-the-art results on the most relevant metrics.

\section{Method}
\label{sec:method}
\subsection{Overview}

The architecture of the proposed method: OMAE which stands for object centric memory-guided auto-encoder is displayed in Fig.~\ref{fig:omae_figure}.First, we detect objects and compute optical flow for each frame. 
Next, appearance $x^{app}$ and motion $x^{mo}$ features are extracted for each object. The former are obtained using a pretrained CNN, whereas the latter consist of motion magnitude and angle maps. We denote the input features $\mathcal{X} = \{ (x^{app}_,x^{mo} )_{j}$ \; $; 1\leq j\leq O\}$,  where $O$ is the total number of objects in the training set. Then, the object representations are encoded and fused into a single embedding $h$.
The three encoders have the same structure: two successive blocks, each block is a shallow fully connected network of two layers. The fusion block consists of a single shallow fully connected network of two layers that learns new embeddings from the concatenation of the three encoders' bottlenecks. 
Thus, we obtain a single hidden representation combining motion and appearance, which can be interpreted as an object-level action feature vector. We will use $\mathcal{H}$ to denote the set of these hidden features corresponding to input vectors in $\mathcal{X}$. This action feature is then used as a query to the \emph{memory of normal patterns} to extract similar existing prototypes in the memory module $\mathcal{M} = \{m_{i} \; ,1\leq i\leq N\}$,  where N is the total number of memory items. 
Similar to \cite{mnad} the most similar memory item $m_{k}$ to the query $h$ is defined as the soft nearest neighbor:

\begin{equation*} \label{eq:attention}
 k = \argmax_{1\leq i\leq N} \left(w_{i}\right) \text{;  } 
 w_{i} = \frac{exp(h^{T}m_{i})}{\sum_{j=1}^{N} exp(h^{T}m_{j})} \text{; } 1\leq i\leq N.
\end{equation*} 

After the memory readout step, and following an attention-mechanism strategy, a linear combination of the memory items is computed as $ c = \sum_{j=1}^{N} w_{j}.m_{j}$  and then concatenated to the query $h$ to obtain an augmented hidden representation $z = (c||h)$ that will be used as input to the decoder network. Finally, a single fully connected decoder network learns to reconstruct $x$ given $z$. 
This way, our auto-encoder model is trained to reconstruct object-centric features $\hat{x}$ under two major constraints:

\textit{1. The auto-encoder learns normality patterns (memory items) that allow the reconstruction of both appearance and motion features from a single embedding space.}

\textit{2. The decoder's reconstructive capacity is limited by the set of memory items thus its generalization ability is reduced, which is useful for detecting anomalies (via poor outlier reconstructions). }

During inference, an \emph{anomaly score} is attributed to each object (cf. Section~\ref{subsec:abn_sco}) based on the dissimilarity between $x$ and $\hat{x}$ as well as the dissimilarity between $h$ and its neighbors in the space of prototypical patterns ($(m_{i})_{1\leq i\leq N}$). 

\subsection{Loss functions}

To take into account the above-mentioned learning constraints, we combine different loss terms in our objective function. We incorporate a reconstruction term $\mathcal{L}_{rec}$ to minimize the discrepancy between the input $x$ and its reconstruction and a memory term $\mathcal{L}_{mem}$ to capture normal prototypical patterns observed in the training set.  Hence, the total loss is given by:
$\mathcal{L} = \mathcal{L}_{rec} + \mathcal{L}_{mem}$.

\textbf{Reconstruction Loss: } We constrain the reconstruction to not only be in the Euclidean neighborhood of the input but also to lie on the same spatial direction. The geometrical constraint is applied via a cosine distance loss and controlled via the hyperparameter $\lambda_{cos}$ such that:
\begin{multline*}
\mathcal{L}_{rec} = \left(\left\lVert x^{app} - \hat{x}^{app}\right\rVert_{2} + \left\lVert x^{mo} - \hat{x}^{mo}\right\rVert_{2}\right)+ \\ \lambda_{cos}\times\left(
1-\frac{<x^{app},\hat{x}^{app}>}{\left\lVert x^{app}\right\rVert_{2} \left\lVert \hat{x}^{app} \right\rVert_{2}}\right) 
\end{multline*}

\textbf{Memory Loss:} This loss is obtained as a combination of three terms:
$$\mathcal{L}_{mem} =\lambda_{comp}.\mathcal{L}_{comp}+ \lambda_{tr}.\mathcal{L}_{tr}+\lambda_{OLE}.\mathcal{L}_{OLE}$$

Similarly to \cite{mnad}, discriminative normality action features prototypes are learnt based on nearest neighbor distances within the memory space via a loss that favors compactness of data samples around prototypes: 
\begin{equation*}
\mathcal{L}_{comp} = \sum_{k=1}^{N}\sum_{j\in\mathcal{U}_{k} \;;\;  \mathcal{U}_{k}\neq \text{\O} } \left\lVert h_{j} - m_{k}\right\rVert_{2} 
\end{equation*}
where $\mathcal{U}_{k} \subset \{1,...,O\}$ is the subset of training object indices which have the memory item $m_{k}$ as their first nearest neighbor.

We also use the same triplet loss $\mathcal{L}_{tr}$ introduced in \cite{mnad}. Contrarily to \cite{mnad}, we incorporate a third term $\mathcal{L}_{OLE}$ that adds orthogonality constraints within the memory space. This is achieved through the geometric loss formulation proposed in \cite{ole_Lezama2018} for supervised classification. We adapt the OLE (Orthogonal Low-rank Embedding) loss to our setting by formulating the memory query step as a classification problem: 
\begin{equation*}
\mathcal{L}_{OLE} = \sum_{c=1}^{N} max(\Delta , \left\lVert H_{c} \right\rVert_{*}) - \left\lVert H \right\rVert_{*} 
\end{equation*}

where $\Delta$ is a positive real number that we set to 1 in all our experiments, $H_{c}$ is the sub-matrix of object hidden representations within the batch that are attributed to memory item $m_{c}$.
$\left\lVert . \right\rVert_{*}$ denotes the nuclear norm i.e the sum of matrix singular values. %

\subsection{Inference: Abnormality score} \label{subsec:abn_sco}

At test time, given the $t^{th}$ frame, a set of abnormality scores is denoted as $\mathcal{S}_{t} = \{s_{t}^{j}\ \; ,1\leq j\leq O_{t}\}$, where $O_{t}$ is the number of detected objects in frame $t$. 

Each score $s_{t}^{j}$ is computed as follows: 
\begin{equation*} \label{eq:ano_score}
	s_{t}^{j} = s = 
	\frac{1}{3}\left(s^{rec}_{L_{2}}+s^{rec}_{cos}+s^{mem}\right)
\end{equation*}
where:

 $s^{rec}_{L_{2}} = g\left(\left\lVert x^{app} - \hat{x}^{app}\right\rVert_{2} + \left\lVert x^{mo} - \hat{x}^{mo}\right\rVert_{2}\right)$

 $s^{rec}_{cos} =g\left( (1-\frac{<x^{app},\hat{x}^{app}>}{\left\lVert x^{app}\right\rVert_{2} \left\lVert \hat{x}^{app} \right\rVert_{2}}) +( 1-\frac{<x^{mo},\hat{x}^{mo}>}{\left\lVert x^{mo}\right\rVert_{2} \left\lVert \hat{x}^{mo} \right\rVert_{2}})\right)$
 
$s^{mem} = g\left(1-\frac{<h,m_k>}{\left\lVert h\right\rVert_{2} \left\lVert m_k \right\rVert_{2}}\right)$

with $g(.)$ a normalization function:

$g(\epsilon) = \frac{\epsilon - \epsilon_{min} }{\epsilon_{max} - \epsilon_{min}}$

where $\epsilon_{min}$ and $\epsilon_{max}$ are the lowest and the highest object level scores respectively across the entire video.

\section{Experiments and results}
\label{sec:format}

\begin{table*}[t!]
        \centering
        \resizebox{1.5\columnwidth}{!}{%
		\begin{tabular}{|c|c|c|c|c|c|c|c|c|c|c|}
			\hline
			\multirow{2}{*}{Approach}
			&\multirow{2}{*}{Method}
			&\multicolumn{3}{c|}{UCSD Ped2}&\multicolumn{3}{c|}{ShanghaiTech}&\multicolumn{3}{c|}{Avenue}\\
			\cline{3-11}
			& & \small AUC&\small RBDC&\small TBDC&  \small AUC&\small RBDC&\small TBDC&  \small AUC&\small RBDC&\small TBDC\\
			
			\cline{3-11}
			
			\hline
		    &MemAE \cite{memae}&94.1 &- &-&71.2 &- &-&83.3 &-&-\\
		    frame-level&MNAD \cite{mnad}&97.0 &- &-&72.5 &- &-&88.5 &- &-\\
			&SSMT \cite{ssmt}&92.4 & - & -&83.5 & - & -&86.9  & - & -\\
			\hline
			\hline
			video &StreetScene \cite{streetscene}&88.3 & 62.50 &80.50&- &- &-&72.0 & 35.80 &\textbf{80.90}\\
			patch-level&Siamese \cite{siamese}&94.0 & \underline{74.0} &89.3&- &- &-&87.2 & 41.20 &\underline{78.60}\\
			\hline
			\hline
			frame \& object level&SSMT \cite{ssmt}&\textbf{99.8} & - & -&\textbf{90.2} & - & -&\underline{92.8}  & - & -\\
			\hline
			\hline
			&dummyAE \cite{dummyae}&94.3 & 52.8 & 72.9 &78.7 & 20.7 & 44.5&87.4 & 15.8 & 27.0\\
			object-level&SSMT \cite{ssmt}&\textbf{99.8} & 72.8 & 91.2&\underline{89.3} & - & -&91.9  & - & -\\
			&BAF \cite{baf}&\underline{98.7} &69.23 &\underline{93.15}&82.7 &\underline{41.43} &\underline{78.79}& 92.3 &\underline{65.05} &66.85\\
			\cline{2-11}
			&OMAE (ours) &96.46&\textbf{80.07}&\textbf{95.39}&79.18&\textbf{51.51}&\textbf{82.19}&93.56*&\textbf{75.83}&70.02\\
			
			\hline
			
		\end{tabular}%
		}
	\caption{Comparison with the state-of-the-art methods (\%). Best results in bold and second best results are underlined. * We use the annotations provided by \cite{streetscene} for Avenue as mentioned in 3.1}
	\label{tab:benchmark}
	
\end{table*}

\textbf{Datasets.}
Several benchmarks had been proposed for evaluating anomaly detection methods \cite{ped,avenue,shanghaitech,ucfcrime,liu,streetscene,adoc},
In order to compare our methods with existing approaches, we performed experiments on the most common datasets : 
\textbf{UCSD ped2}~\cite{ped} includes 16 training and 12 testing videos of resolution 240x360. Anomalous events include riding a bike and driving a vehicle on a sidewalk.
\textbf{CUHK Avenue}~\cite{avenue} consists of 16 training and 21 test videos of resolution 360x640  with abnormal events such as running, walking towards the camera, or throwing papers. We use the annotations provided by \cite{streetscene}.
\textbf{ShanghaiTech}~\cite{shanghaitech} contains 330 training and 107 testing videos of resolution 480x856 with 13 different scenes. Each scene has a different background or camera angle. Abnormal events include jumping, running, or stalking on a sidewalk.
\vspace{1mm}

\noindent\textbf{Evaluation metrics.} Since we focus on spatio-temporal 
anomaly detection, we
adopt the Region-Based Detection Criterion (RBDC)
and the Track-Based Detection Criterion (TBDC) 
metrics
introduced in
\cite{streetscene} as an alternative to the flawed pixel-level AUC metric. 
We also report the Area Under of the ROC Curve (AUC) 
obtained with respect to the frame-level ground-truth annotations. 
Yet, it gives only a global frame score and, therefore, doesn't reflect the model capacity to localize anomalies. Emphasis is given to the parts of the ROC curve  where false positive rate is too high for a practical use \cite{lobo2008auc}. Hence, AUC is the least relevant metric in our study.

\vspace{1mm}

\noindent\textbf{Parameters and implementation details.} \label{subsec:implementation} The first step of our framework is to perform object detection using Yolov3 \cite{yolov3} pretrained on COCO dataset as in \cite{ssmt,baf,dummyae}. We set the detection confidence to 0.7 on ShanghaiTech and Avenue for both the training and testing sets. Since the image resolution of UCSD ped2 is lower, we reduced the threshold to 0.5. We used ResNet101 to precompute appearance features of detected objects and 
Farneback's algorithm to compute optical flows \cite{farneback2003two}. Similarly to \cite{ssmt},  we trained the network for 30 epochs on each dataset using Adam optimizer with a learning rate of $10^{-3}$. We use a batch size of 256 object-level action features for the smallest dataset (UCSD ped2) and 512 for ShanghaiTech and Avenue. 
For all experiments, we set $\lambda_{cos}$ = 0.1 so that the cosine term has a similar order of magnitude as the $L_2$ reconstruction term;
as well as fixed empirical values for the number of memory items and the memory loss weights: $N = 40$, $\lambda_{comp}$ = 1.6 , $\lambda_{tr}$ = 0.2, $\lambda_{OLE}$ = 0.3. 
The epoch achieving the lowest loss value in training is used for inference. As a post-processing step and similarly to \cite{baf}, we apply a spatio-temporal mean filtering to smooth object level scores and a Gaussian filter at the frame level. In the case of Avenue dataset,
scale change is taken into account by an anomaly score adjustment: the anomaly score of an object is multiplied by its bounding box width. 
This post-processing allows an increase of 26\% in RBDC and 2\% in TBDC
without degrading AUC.

\vspace{1mm}

\noindent\textbf{Ablation study.} We conduct an ablation study on ShanghaiTech dataset to assess the importance of each component in our framework. The corresponding results are shown in Table \ref{tab:ablative}. We can see that both appearance and motion features are necessary to model usual actions to better detect anomalies. Indeed, the baseline model takes only appearance features as input and performs lower than the current state of the art on all metrics. 
Including the motion information through optical flow improves the frame-AUC by 8.97\% and significantly increases RBDC and TBDC by 15.83\% and 32.37\% respectively allowing OMAE to outperform the current state of the art on these metrics. 
We also note that including the cosine similarity together with MSE or MAE in the reconstruction loss as well as in the abnormality scores improves all metrics significantly, showing the importance of the orthogonality constraints.

\begin{table}
        \centering
        \resizebox{0.9\columnwidth}{!}{%
		\begin{tabular}{|c|c|c|c|c|c|c|c|}
			\hline
			\multirow{2}{*}{Features} & \multicolumn{2}{c|}{\small Reconstruction} & %
			\multicolumn{2}{c|}{Anomaly}& \multicolumn{3}{c|}{Evaluation} \\
			 & \multicolumn{2}{c|}{loss \tiny (AE)}  &\multicolumn{2}{c|}{score}
			&\multicolumn{3}{c|}{metrics}\\
			\cline{2-8} & \small MSE &\small  COS &\small MAE&\small COS
			&\small AUC&\small RBDC&\small TBDC\\
			\hline
			appearance&\checkmark & \checkmark&\checkmark&\checkmark \tiny (AE+Mem)& 70.21& 35.68 &  49.82\\
			\hline
			& \checkmark & - &\checkmark& - &70.48&35.15&69.16   \\
			& \checkmark & -&- & \checkmark\tiny (Mem)& 71.61& 27.72 & 57.70   \\
			appearance& \checkmark & - & -&\checkmark\tiny(AE)&76.66&43.02& 68.13  \\
			+ & \checkmark & -&\checkmark&\checkmark\tiny (AE+Mem)& 75.98& 39.53 & 67.49   \\
			\cline{2-8}
			motion& \checkmark & \checkmark&\checkmark &- &70.71&35.96&71.27  \\
			& \checkmark & \checkmark&- & \checkmark\tiny (Mem)& 69.50& 31.18 & 74.47   \\
			& \checkmark & \checkmark&-&\checkmark \tiny (AE) &76.68&43.99&69.72\\
			& \checkmark & \checkmark&\checkmark&\checkmark\tiny (AE+Mem)& 77.81	&49.37	&\textbf{83.61}	 \\
			\hline
			+ 3D smoothing & \checkmark & \checkmark&\checkmark&\checkmark\tiny (AE+Mem)& \textbf{79.18}& \textbf{51.51} & 82.19 \\
		
			\hline
		\end{tabular}%
		}
		\caption{Frame-AUC, RBDC and TBDC scores (in \%) obtained on ShanghaiTech by making gradual design changes to the baseline method, until the final framework.
		(\textit{AE} stands for the auto-encoder component; \textit{Mem} for the nearest-neighbor memory item.)
		}
		\label{tab:ablative}
	
\end{table}

\vspace{1mm}

\noindent\textbf{Comparison with state of the art.} In Table \ref{tab:benchmark}, we present our results in comparison with the state-of-the-art methods on the 3 benchmark datasets. Our framework significantly outperforms current state-of-the-art methods on the most relevant evaluation metrics RBDC and TBDC which quantifies the model's ability to localize anomalies spatially and to track them temporally. It's important to highlight that, given a single dataset, there is not a single best method that outperforms the other approaches on the three metrics. On ShanghaiTech dataset, we outperform best previous work by a margin of 10.08\% on RBDC and 3.4\% on TBDC while remaining competitive with respect to other object-centric approaches in terms of frame-level AUC. Our model also reaches new state-of-the-art performances in terms of RBDC and TBDC on UCSD ped2 with significant margins of  6.07\% and 2.24\% respectively.  On Avenue, our model improves the current state of the art by 10,78\% on the RBDC metric and outperforms the recent object-centric approach \cite{baf} in terms of TBDC by a margin of 3.17\% while remaining competitive on the AUC metric. Advantageously, inference time takes only 20ms for a batch of 256 precomputed object action features. The pre-processing time required for computing those features is the following: optical flow
extraction (170ms), object detection (50ms) and feature extraction (40ms). In addition, training \footnote[3]{This publication was made possible by the use of the FactoryIA supercomputer, financially supported by the Ile-de-France Regional Council} is much faster than in~\cite{mnad}: only 70 min on ShanghaiTech, 9 min on Avenue and 4 min on UCSD ped2 with a single NVIDIA TITAN X (PASCAL) GPU.

\section{Conclusions}
In this work we introduced OMAE, an object-centric VAD framework that uses a memory module for object-level appearance and motion features  with a new abnormality scoring strategy based on cosine distance. In our experiments, OMAE reaches superior results on localization and tracking metrics while remaining competitive on the frame-level AUC. This shows the effectiveness of our approach to localize anomalies better than the current state of the art.

\bibliographystyle{IEEEbib}
\bibliography{main}

\end{document}